\documentclass[journal]{IEEEtran}
\usepackage{graphicx}
\usepackage{color}
\usepackage{amsmath}
\usepackage{amssymb}
\usepackage{subfigure}
\usepackage{tabularx}
\usepackage{times}
\usepackage{xtab}
\usepackage{setspace}
\usepackage{bm}
\usepackage{relsize}
\usepackage{accents}
\usepackage{hyperref}
\usepackage{cite}

\DeclareFontFamily{OT1}{pzc}{}
\DeclareFontShape{OT1}{pzc}{m}{it}{<-> s * [1.30] pzcmi7t}{}
\DeclareMathAlphabet{\mathpzc}{OT1}{pzc}{m}{it}

\allowdisplaybreaks

\title{Load Disaggregation Based on\\Aided Linear Integer Programming}

\author{Md.~Zulfiquar~Ali~Bhotto, 
Stephen Makonin, 
        and Ivan V. Baji\'{c}
\thanks{Manuscript received Mar.~2016, revised Jun. and Jul. 2016. This work was supported by the Natural Sciences and Engineering Research Council of Canada under Grant EGP-486210-15. }
\thanks{ The authors are with the School of Engineering Science, Simon Fraser University, Burnaby, BC
          V5A 1S6, Canada (e-mail: mbhotto@sfu.ca, smakonin@sfu.ca, ibajic@ensc.sfu.ca).}
}

\begin{document}

\date{}




\maketitle

\begin{abstract}
Load disaggregation based on aided linear integer programming (ALIP) is proposed. We start with a conventional linear integer programming (IP) based disaggregation and enhance it in several ways. The enhancements include additional constraints, correction based on a state diagram, median filtering, and linear programming-based refinement. With the aid of these enhancements, the performance of IP-based disaggregation is significantly improved. The proposed ALIP system relies only on the instantaneous load samples instead of waveform signatures, and hence works well on low-frequency data. Experimental results show that the proposed ALIP system performs better than conventional IP-based load disaggregation.
\end{abstract}

\section{Introduction}
\label{sec:secII}

Load disaggregation or non-intrusive load monitoring (NILM) is the process of finding out how much each appliance within a household is consuming when only the aggregate current or power reading is available~\cite{makonin13}. Such readings are now available through smart meters, which have been, or are being, installed by most power utilities. In addition to determining appliance consumption patterns, NILM could help  balance different loads within a power network~\cite{zdraveskidynamic} by predicting demand without the use of additional sensors.

Recent disaggregation methods make use of machine learning approaches such as
clustering~\cite{koutitas16},
fuzzy systems~\cite{ducange14},
and hidden Markov models~\cite{kolter2011,johnson13,zeifman12,egarter15,makonin15}. 
Such methods might lead to practical solutions when large and sufficiently representative datasets become available for training, which is still not the case. 
The ultimate goal for NILM is to enable disaggregation without the need for supervised learning. There has been some recent progress in this area~\cite{zhao2016training,egarter2015autonomous}, although the accuracy of such methods is still low.
Alternative approaches such as combinatorial optimization or integer programming (IP) have been much less explored.
Two notable earlier works on IP-based disaggregation include Egarter~\textit{et~al.}~\cite{egarter13} and Suzuki~\textit{et~al.}~\cite{sujuki2008}. Egarter~\textit{et~al.} formulated disaggregation as a modified knapsack problem and proposed a solution using an evolutionary algorithm.
From a practical standpoint, the drawback here is that evolutionary algorithms potentially have a long execution time, and their stochastic nature may lead to different solutions in different runs, even on the same data. Our goal here is to develop a simpler and more principled approach that gives repeatable results.

The other IP-based disaggregation proposal was by Suzuki~\textit{et~al.}~\cite{sujuki2008} in 2008 using high-frequency sampled current readings.
We extract the IP portion of~\cite{sujuki2008} (without the load signature part) and enhance it in multiple ways to improve disaggregation accuracy. The enhancements include additional constraints, correction based on a state diagram, median filtering, and linear programming-based refinement.Our method works on low-frequency data, which is a more realistic solution for current smart meters that usually report power readings at intervals of 8--15 seconds\footnote{For example, Rainforest Automation's EMU2 device polls the smart meter at 15 second intervals, while their Eagle product polls at 8 seconds.  In the UK it is mandated to 10 seconds\cite[p. 78]{smimp}.}.

The remainder of the paper is organized as follows. We mathematically define the problem of load disaggregation as a  mixed-integer linear IP problem in~Section~\ref{sec:sec2}. We propose several enhancements to IP-based disaggregation in~Section~\ref{sec:sec4}, which is our main contribution. A number of experimental results with a comparison to previous works are reported in~Section~\ref{sec:sec5}, followed by conclusions in~Section~\ref{sec:sec6}.

\section{Load Disaggregation as Integer Programming}
\label{sec:sec2}

Consider a household with $n$ appliances, where the $i$-th appliance ($i=1,2,\ldots,n$) has $l_i$ non-OFF states. For example, a conventional light bulb would have one non-OFF state.
Vector $\mathbf{r}_i\in\mathbb{R}^{l_i}$ contains the  voltampere (VA) ratings of the $l_i$ non-OFF states of the $i$-th appliance. Let $m=\sum_{i=1}^nl_i$. With this we construct vector $\mathbf{r}=[\mathbf{r}_1; \mathbf{r}_2;\cdots;\mathbf{r}_n]$
that contains all VA (or Watt) ratings of all $n$ appliances.  For the $k$-th time instant, the indicator of each non-OFF state  is stored in a vector $\mathbf{b}_k$ as
\begin{equation}
\label{eqn:eqn1}
\mathbf{b}_k[i]\in\{1,\ 0\}
 \mathrm{\ for\ } i=1,2,\ldots,m,
 \end{equation}
with $1$ denoting that the particular state is active and $0$ denoting the state is inactive.

At the $k$-th time instant, the smart meter yields the total VA reading $z_k$, which is the sum of all power drawn by $n$ appliances at that time.
This can be expressed as
\begin{eqnarray}
\label{eqn:eqn2}
\mathbf{s}_k&=& \mathbf{F}~\texttt{diag}(\mathbf{b}_k)\mathbf{r}\\
\label{eqn:eqn3}
z_k&=&\mathbf{h}^t\mathbf{s}_k
\end{eqnarray}
where $\mathbf{h}=[1;\ 1;\ \cdots;\ 1]\in\mathbb{R}^{n}$
and vector $\mathbf{s}_k\in\mathbb{R}^{n}$ contains VA draws of each appliance that is turned ON. The binary
matrix $\mathbf{F}\in\mathbb{R}^{n\times m}$ in (\ref{eqn:eqn2}) is a block diagonal matrix given as
\[\mathbf{F}=\texttt{diag}(\mathbf{1}_1^t,\ \mathbf{1}_2^t,\ \ldots, \mathbf{1}_n^t)\]
where $\mathbf{1}_i\in\mathbb{R}^{l_i}$ is a unity vector for $i=1,2,\ldots, n$. 

The objective of load disaggregation is to find which appliance states are active at the $k$-th time instant. Specifically, the goal is to find $\mathbf{b}_k$ in (\ref{eqn:eqn2}) by using the known quantities $z_k$, $\mathbf{F}$, and $\mathbf{r}$. Similarly to~\cite{sujuki2008}, load disaggregation can be formulated as an integer programming (IP) problem
\begin{equation}
\label{eqn:eqn4}
\renewcommand{\arraystretch}{0.3}
\begin{array}{c c}
\mathrm{minimize} & \left(z_k-
\mathbf{h}^t\mathbf{F}~\texttt{diag}(\mathbf{b}_k)~\mathbf{r}\right)^2.\\
\mathbf{b}_k &
\end{array}
\end{equation}
Like any IP, this problem can be solved by exhaustive search over all possibilities for $\mathbf{b}_k$, however this approach can be prohibitive even for a moderate number of appliances.
The alternative is to explore more efficient IP solvers~\cite{aawl07}.

Before proceeding to enhancements, we reformulate~(\ref{eqn:eqn4}) as a mixed-integer linear IP. Let $q_j=\sum_{i=1}^{j-1}l_i$. Since any appliance must be in exactly one state at any given time and vector $\mathbf{b}_k$ is an indicator for non-OFF states, we can formulate this constraint as
\[\mathbf{b}_k[q_j+1]+\mathbf{b}_k[q_j+2]+\cdots+\mathbf{b}_k[q_j+l_j]\le 1\]
for $j=1,2,...,n$. Let $\mathbf{v}=\mathbf{r}\odot(\mathbf{F}^t\mathbf{h})$ where $\odot$ denotes the element-wise product. Further, let $\check{\mathbf{v}}$ $=$ $[0;\ \mathbf{v}]$, $\mathbf{u}_1 $ $=$ $[1;\ \mathbf{0}]$,  $\mathbf{f}  = [1;\ \mathbf{0}]$, and $\mathbf{x}_k=[\delta;\ \mathbf{b}_k]$, where $\delta$ is an auxiliary real variable. We can now rewrite the quadratic IP in~(\ref{eqn:eqn4}) as a mixed-integer \emph{linear} IP
\begin{equation}
\label{eqn:eqn9}
\renewcommand{\arraystretch}{0.3}
\begin{array}{c c}
\mathrm{minimize} & \mathbf{f}^t\mathbf{x}_k\\
\mathbf{x}_k
\end{array}
\end{equation}
subject to
\begin{eqnarray}
\label{eqn:eqn10}
\mathbf{A}\mathbf{x}_k&\le&\mathbf{e}\\
\label{eqn:eqn11}
\mathbf{x}_k[i]&\in&\{1, 0\} \textrm{\ for\ } i=2,3,\ldots,m+1
\end{eqnarray}
where matrix $\mathbf{A}$ and  vector $\mathbf{e}$ in (\ref{eqn:eqn10}) are given by
$
\mathbf{A}$ $=$
$[-(\check{\mathbf{v}} +\mathbf{u}_1 )^t;$ 
$(\check{\mathbf{v}} -\mathbf{u}_1 )^t;$ 
$\check{\mathbf{F}}]$ and
$\mathbf{e}$ $=$
$[-z_k;$ $z_k;$ $\mathbf{1}]$
and the rows of the binary matrix $\check{\mathbf{F}}$ are copies of the rows of matrix $\mathbf{F}$ that have more than one nonzero element. 

The solution of (\ref{eqn:eqn9})--(\ref{eqn:eqn11}) leads to correct disaggregation only if the elements in $\mathbf{r}$ are not binary combinations (linear combinations with coefficients $0$ or $1$) of each other, and the deviations from the steady-state power/current draw during transients do not overlap with the steady-state draws of other appliances, or their combinations. In practice, these criteria are rarely met, so disaggregation based on (\ref{eqn:eqn9})--(\ref{eqn:eqn11}) would yield unsatisfactory results. This can also be true for an appliance with an infrequently occurring state with a high rating, since an undetected state with high rating would significantly reduce the accuracy score (Section~\ref{sec:sec5}).
In the next section we introduce several enhancements that are meant to overcome the aforementioned limitations of the above IP disaggregation.

\section{Aided Linear IP for Load Disaggregation}
\label{sec:sec4}

The proposed aided linear IP (ALIP) for load disaggregation incorporates several enhancements to the IP given in (\ref{eqn:eqn9})--(\ref{eqn:eqn11}), each of which is discussed next.

\subsection{Additional Constraints}
The first enhancement involves additional constraints that help resolve ambiguities related to the possible non-uniqueness of the solution to the IP.
First, consider appliances like refrigerator, surveillance camera, smoke detector,  heat pump, etc., that happen to  switch between the ``sleep mode'' and one or more higher-power states. In other words, these appliances always draw some power. We can incorporate this additional information as an equality constraint to be added to (\ref{eqn:eqn9})--(\ref{eqn:eqn11}),
\begin{equation}
\label{eqn:eqn14}
\mathbf{A}_{eq}\mathbf{x}_k=\mathbf{1}
\end{equation}
where each row in the binary matrix $\mathbf{A}_{eq}$ has unity elements only in those positions that correspond to the states of those appliances in vector $\mathbf{x}_k$ that remain turned ON at all time.

Next, consider the scenario where the rating of a given state of an appliance is equal to the sum of the ratings of some of the states of other appliances. For example, an appliance H1 could have a VA rating $300$ in one of its states, and appliances H2 and H3 could have VA ratings of $100$ and $200$, respectively, in some of their states. Then a reading $z_k=300$ could be interpreted in two ways - either H1 is ON, or H2 and H3 are simultaneously ON. To break such ties, we assume the minimum number of appliances is ON at any given time - a heuristic that does not always hold, but turns out to be surprisingly good based on empirical evidence in existing datasets.
This can be incorporated into the IP by using  an additional row in the binary $\mathbf{A}$ and an additional $1$ in $\mathbf{e}$ in (\ref{eqn:eqn10}). The additional row in $\mathbf{A}$ would have unity elements only in those positions that correspond to the states where the ratings become binary combinations of each other.

Finally, consider the scenario where the rating of a given appliance (say H1) is equal to the amount by which the steady-state rating of another appliance (say H2) differs from its transient VA measurement. The transient reading may cause the IP solver to declare both H1 and H2 as ON, even though only H2 is in the ON transient.
Such situation can also be avoided by using  an additional row in the binary matrix $\mathbf{A}$ and an additional $1$ in $\mathbf{e}$ in (\ref{eqn:eqn10}), where the additional row in $\mathbf{A}$ has unity elements only in those positions that correspond to the states whose combinations produce a transient rating of another appliance.

\subsection{Correction Based on State Transition Diagrams}
Many appliances operate as finite state machines and their possible state transitions can be described by a state transition diagram (STD). For example, the fridge (FRG) appliance from the dataset in~\cite{makonin16} has the STD shown in Fig.~\ref{fig:fig1}. This offers the possibility to correct the output of an IP solver if it happens to violate the STD. For example, if FRG was in state $s_1$ at time $k-1$, then at time $k$ it can only be in $s_1$ or $s_2$. If the IP solver output suggests otherwise, we know there must be an error, and can therefore select either $s_1$ or $s_2$, depending on which of the two options yields lower cost $\mathbf{f}^t\mathbf{x}_k$ in~(\ref{eqn:eqn9}). The same type of correction can also be applied backwards (for example, the only way to get to $s_3$ is either from $s_2$ or $s_3$), although we did not incorporate such processing in our experiments.

\subsection{Median Filtering}

\begin{figure}
  \centering
  \includegraphics[width=0.55\columnwidth]{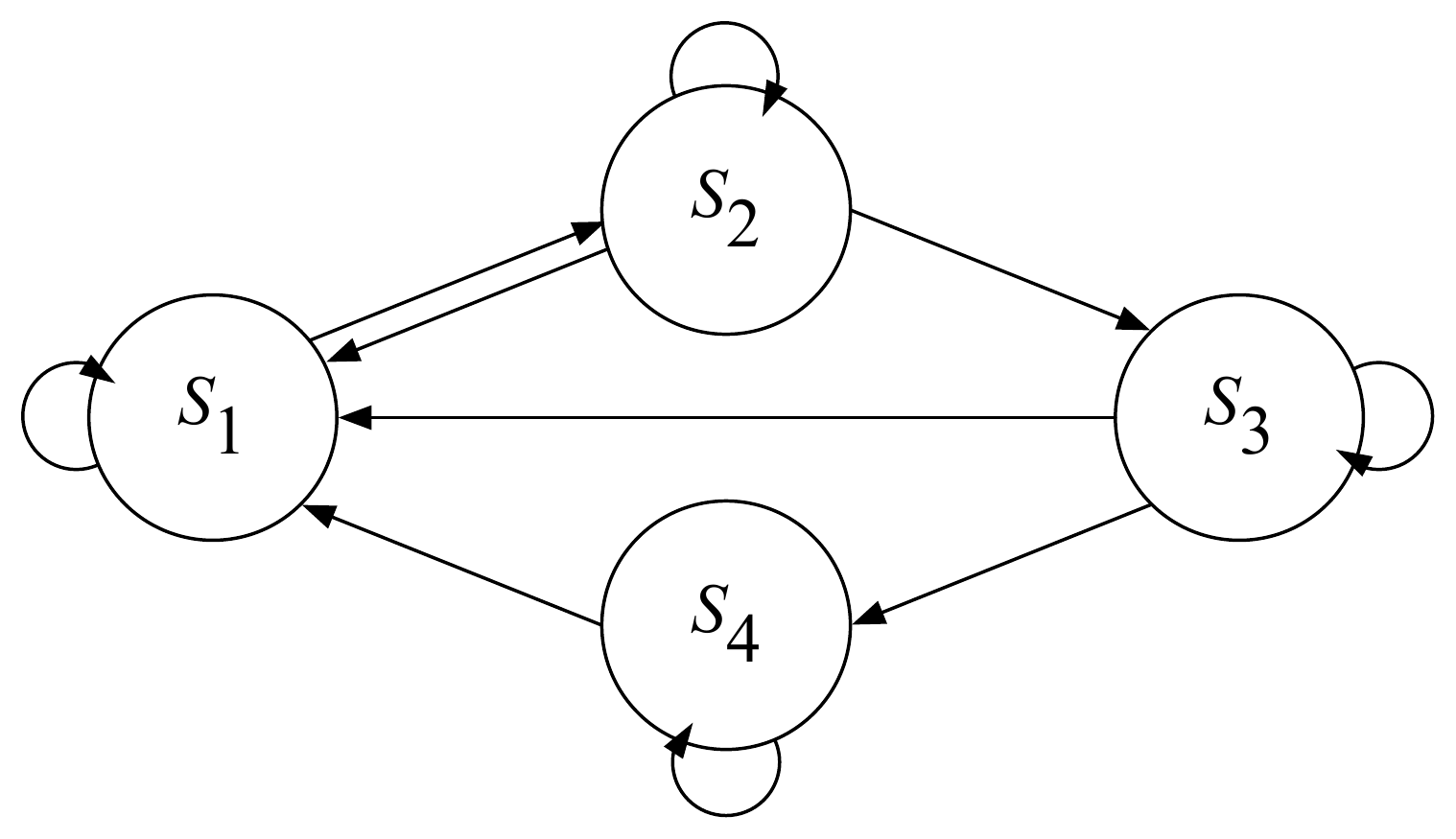}
    \vspace{-0.1 in}
  \caption{State transition diagram (STD) of FRG from~\cite{makonin16}.}\label{fig:fig1}
      \vspace{-0.1 in}
\end{figure}

Median filtering can help filter out implausible events such as frequent switching between states, which may occur in the IP solver output if the ratings of a particular appliance are much smaller than the total reading $z_k$. Consider the appliance B1E from~\cite{makonin16}, which has a fully-connected 2-state diagram (Fig.~\ref{fig:fig1} with states $s_3$ and $s_4$ deleted). Although any transition between these two states is possible, it is implausible that the appliance changes state at each sampling instant; we expect it to stay in a state for at least a few sampling instants.
To enforce this, we apply the correction rule
\begin{equation}
\label{eqn:eqn17}
\hat{s}_{k-L}=
\begin{cases}
s_1  \textrm{\ if\ } \hat{s}_{k-L}=s_2 \textrm{\ and\ } \textrm{med}(\hat{s}_{k},
\hat{s}_{k-1},\ldots,\hat{s}_{k-L})=s_1 \\
s_2 \textrm{\ if\ } \hat{s}_{k-L}=s_1 \textrm{\ and\ } \textrm{med}(\hat{s}_{k},
\hat{s}_{k-1},\ldots,\hat{s}_{k-L})=s_2 \\
\end{cases}
\end{equation}
for $k>L$ in the solution obtained by the IP solver. In other words, the state estimate at time $k-L$ (i.e., $\hat{s}_{k-L}$) is corrected based on the current state estimate $\hat{s}_k$ and the corrected state does not affect the subsequent median filter outputs. Analogous corrections are applied to all states of all appliances.

\begin{figure}
  \centering
  \includegraphics[width=.9\columnwidth]{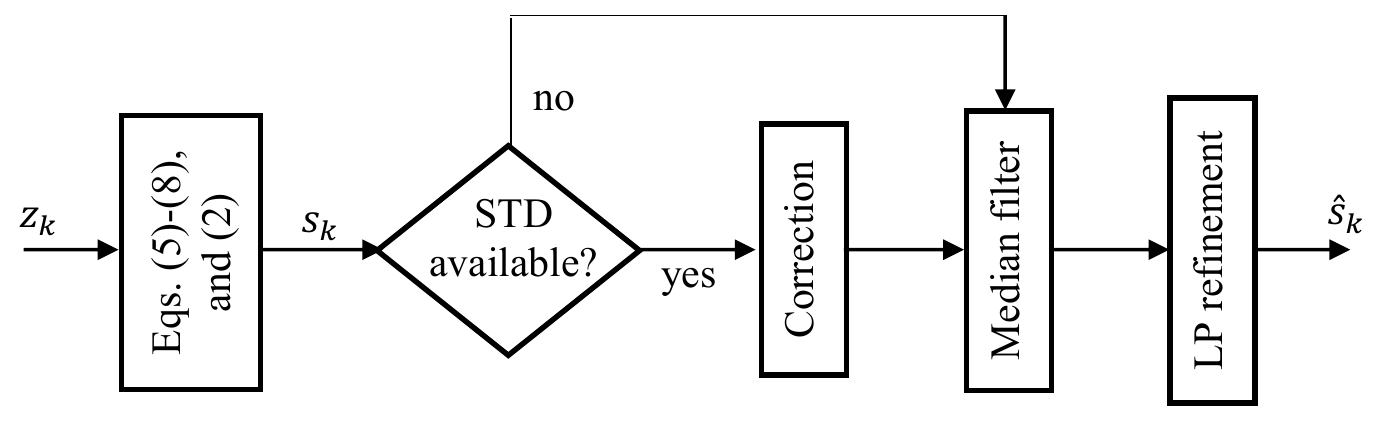}
  \vspace{-0.1 in}
  \caption{ALIP flowchart. }\label{fig:fig2}
  \vspace{-0.1 in}
\end{figure}

\subsection{Linear Programming-Based Refinement}
As mentioned before, vector $\mathbf{r}$ contains steady-state appliance ratings, which could be obtained from appliance data sheets or measurements. However, VA (or Watt) values for transients between states are usually much more difficult to obtain, and even if this were possible, incorporating transient state ratings into the model would tremendously increase the model size (i.e., number of states) $m$. Yet, if the sampling instance happens to catch a transient of one or more appliances, it could lead to incorrect solution of the IP. For this reason, we develop a method to refine the IP solution using only the minimum and maximum transient rating of each appliance, which is easier to obtain, either from the data sheet or measurement.

Let $\mathbf{r}_{\mathrm{min}}$ and $\mathbf{r}_{\mathrm{max}}$ be the vectors of the same size as $\mathbf{r}$ that contain, respectively, the minimum and maximum transient ratings for each state of each appliance. Let vector $\mathbf{p}_1$ contain indices of $\mathbf{r}$ for which $\mathbf{r}_{\mathrm{min}}(\mathbf{p}_1)$ $=$ $\mathbf{r}_{\mathrm{max}}(\mathbf{p}_1)$ $=$ $\mathbf{r}(\mathbf{p}_1)$. Such states do not exhibit transient behavior. Let the vector $\mathbf{p}_2$ contain the indices of other, potentially transient, states, that have been declared active by the IP solver (i.e., the corresponding value in $\mathbf{b}_k$ is $1$). If $\mathbf{p}_2$ is non-empty, then the current measurement may contain transient states and the solution given by the IP solver needs to be refined.

Let $\bm{\mathbin{h}}$ $=$ $\mathbf{F}^t\mathbf{h}$ and $\mathbf{w}_k = \mathbf{b}_k\odot\mathbf{r}$. To refine the solution, we solve the following problem
\begin{equation}
\label{eqn:eqn18}
\renewcommand{\arraystretch}{0.3}
\begin{array}{c c}
\mathrm{minimize} & \left(z_k-
\bm{\mathbin{h}}^t\mathbf{y}_k\right)^2\\
\mathbf{y}_k &
\end{array}
\end{equation}
subject to
\begin{eqnarray}
\label{eqn:eqn19}
&~&\mathbf{y}_k(\mathbf{p}_1)=\mathbf{w}_k(
\mathbf{p}_1)\\
\label{eqn:eqn20}
&~&\mathbf{r}_{\mathrm{min}}
(\mathbf{p}_2)\le\mathbf{y}_k(\mathbf{p}_2)
\le\mathbf{r}_{\mathrm{max}}(\mathbf{p}_2)
\end{eqnarray}
Constraints (\ref{eqn:eqn19})--(\ref{eqn:eqn20}) force the non-transient states to match the steady-state ratings in $\mathbf{r}$ while requiring potentially transient states to be between the corresponding minimum and maximum.

The cost function can be simplified by subtracting out the steady-state portion of the measurement, $\check{z}_k$ $=$ $z_k$ $-$ $\bm{\mathbin{h}}(\mathbf{p}_1)^t\mathbf{w}_k(\mathbf{p}_1)$, and focusing on the transient portion of $\mathbf{y}_k$, i.e., $\check{\mathbf{y}}_k = \mathbf{y}_k(\mathbf{p}_2)$. Then, applying a similar procedure as in Section~\ref{sec:sec2}, the problem can be converted to a linear programming problem. With
$\check{\mathbf{x}}=[\delta;\ \check{\mathbf{y}}_k]$,  $\check{\bm{\mathbin{h}}}$ $=$ $[0;\ \mathbf{1}]$, $\mathbf{u}_1 $ $=$ $[1;\ \mathbf{0}]$,  and $\mathbf{f}  = [1;\ \mathbf{0}]$, the problem becomes
\begin{equation}
\label{eqn:eqn23}
\renewcommand{\arraystretch}{0.3}
\begin{array}{c c}
\mathrm{minimize} & \mathbf{f} ^t\check{\mathbf{x}}\\
\check{\mathbf{x}}
\end{array}
\end{equation}
subject to
\begin{equation}
\label{eqn:eqn24}
\mathbf{A}\check{\mathbf{x}}\le\mathbf{e}
\end{equation}
where matrix $\mathbf{A}$ and  vector $\mathbf{e}$ in (\ref{eqn:eqn24}) are given by
$
\mathbf{A}$ $=$ 
$[
-(\check{\bm{\mathbin{h}}} +\mathbf{u}_1 )^t;$ $ 
(\check{\bm{\mathbin{h}}} -\mathbf{u}_1 )^t;$
$[\mathbf{0} $  $-\mathbf{I}];$ $[\mathbf{0}$ $\mathbf{I};]]$
and
$
\mathbf{e}=
\begin{bmatrix}
-\check{z}_k;&
\check{z}_k;&
-\check{\mathbf{r}}_l;&
\check{\mathbf{r}}_u
\end{bmatrix}
,$
respectively. 

The flowchart of ALIP is shown in Fig.~\ref{fig:fig2}.
Further steps such as time-of-day probabilities can be  incorporated in order to further improve disaggregation accuracy.


\section{Experimental Results}
\label{sec:sec5}
We compare the performance of our ALIP method with the IP-based disaggregation from~\cite{sujuki2008} in terms of two accuracy measures~\cite{makonin_popowich15}: per appliance accuracy
 \[\mathrm{AC}_i=1-\frac{\sum_{k=1}^N
|\mathbf{s}_k[i]-\hat{\mathbf{s}}_k[i]|}{2
\sum_{k=1}^N|\mathbf{s}_k[i]|}\]
and overall accuracy
\[\mathrm{ACC}=1-\frac{\sum_{k=1}^N\sum_{i=1}^n
|\mathbf{s}_k[i]-\hat{\mathbf{s}}_k[i]|}{2
\sum_{k=1}^N\sum_{i=1}^n|\mathbf{s}_k[i]|}\]
where $\mathbf{s}_k[i]$ is the ground-truth rating of the $i$-th appliance at time index $k$ from the dataset and $\hat{\mathbf{s}}_k[i]$ is its estimate obtained by disaggregation. In all experiments, the steady-state ratings of each appliance were determined empirically from the datasets from the power consumption of each appliance separately. The maximum and minimum transient VA or Watt values for the ALIP disaggregator were also determined empirically. The aggregate VA or Watt totals for the appliances used in the experiments were computed from aggregated individual appliance readings. For ALIP, the enhancements were applied in the order depicted in the flowchart in Fig.~\ref{fig:fig2}. All ratings and parameters used in the experiments can be found in the Matlab code\footnote{\url{http://www.sfu.ca/~ibajic/software/NILM-TCAS.rar}}, which can be used to reproduce all the results.

\begin{figure}
  \centering
  \includegraphics[width=0.80\columnwidth]{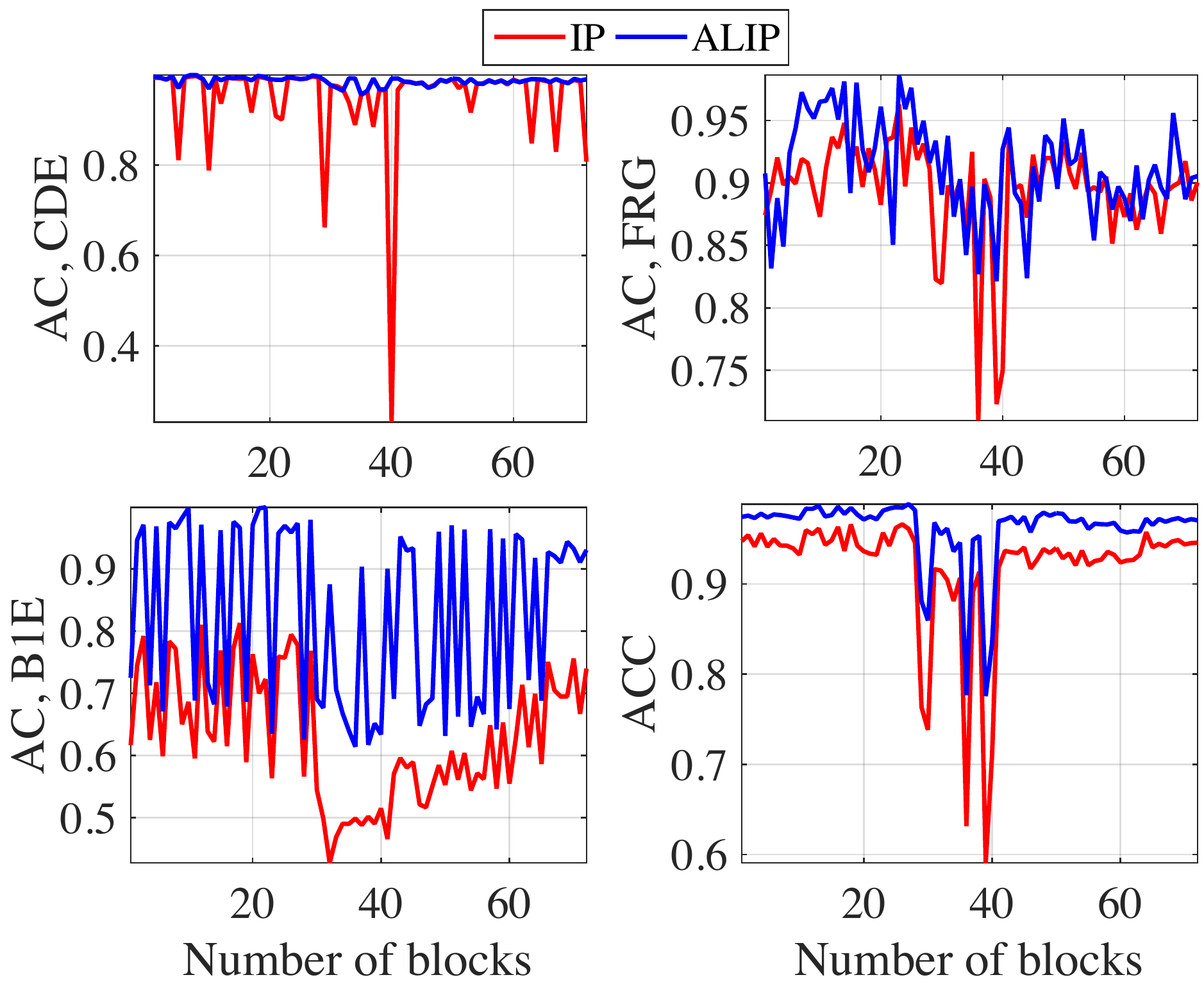}
      \vspace{-0.15 in}
  \caption{AC and ACC plots for Experiment 1. }\label{fig:fig3}
      \vspace{-0.2 in}
\end{figure}

In Experiment 1 we used  $n=4$ appliances (CDE, FRG, HPE, and B1E) from the dataset in~\cite{makonin16}. The number of states considered for the CDE, FRG, HPE, and B1E appliances were $3$, $4$, $4$, and $2$, respectively. The total number of samples considered was $72\times5040$, which we partitioned into 72 blocks of data each containing $N=5040$ samples. This number of samples covers $9$ months worth of readings.  We computed AC and ACC for each block. The AC curves obtained by using the IP and proposed ALIP disaggregators for CDE, FRG, and B1E appliances are illustrated in Fig.~\ref{fig:fig3}, along with the ACC curves. We have not illustrated AC curves for HPE, since both disaggregators produced very similar curves for this appliance.

It is seen from the AC, CDE plots (top-left in Fig.~\ref{fig:fig3}) that ALIP  performs better than IP  consistently in all blocks. The same is true for B1E (bottom-left in Fig.~\ref{fig:fig3}). From the AC, FRG plots it is seen that the ALIP performs considerably better than IP in many blocks, while IP  performs marginally better in some blocks. This is because FRG has occasional impulsive VA readings which get filtered out by ALIP but not by IP. Overall, however, ALIP disaggregates FRG better than IP.
 From the ACC plots (bottom-right of Fig.~\ref{fig:fig3}), it can be seen the ALIP disaggregator is overall more accurate than the IP disaggregator - usually by 5--8\%, and in some cases by up to 20\%.  The AC values obtained over the whole 9 months worth of data are given in Table~\ref{tbl:allres}, while the overall ACC measure for this and other experiments is given in Table~\ref{tbl:overallres}. As seen in these tables, ALIP outperforms IP on each appliance, as well as overall.

\begin{table}[t]
  \centering
  \caption{Results for Exp. 1--7}
  \footnotesize
  \begin{tabular}{l|l|c|l|r|r}
  \hline
  \hline
  Exp. & Dataset & ID & Appliance & IP & ALIP \\
  \hline

  1 & AMPds   & CDE & Clothes Dryer       & 0.986 & 0.987 \\
  ~ & ~       & FRE & Furnace Fan         & 0.891 & 0.909 \\
  ~ & ~       & HPE & Heat Pump           & 0.955 & 0.957 \\
  ~ & ~       & B1E & Bedroom             & 0.600 & 0.757 \\
  \hline

  2 & REDD    & OVN &  Oven               & 0.65  & 0.63  \\
  ~ & House 1 & FRG &  Refrigerator       & 0.79  & 0.85  \\
  ~ & ~       & DSH &  Dishwasher         & 0.88  & 0.92  \\
  ~ & ~       & MIC &  Microwave          & 0.74  & 0.83  \\
  ~ & ~       & DRY &  Clothes Dryer      & 0.64  & 0.78  \\
  ~ & ~       & BTH &  Bathroom GFI       & 0.64  & 0.70  \\
  ~ & ~       & DIF &  Unmetered          & 0.60  & 0.95  \\
  \hline

  3 & REDD    & KTC & Kitchen Outlets     & 0.88  & 0.95  \\
  ~ & House 2 & LTE & Lighting            & 0.91  & 0.96  \\
  ~ & ~       & STV & Stove               & 0.71  & 0.75  \\
  ~ & ~       & MIC & Microwave           & 0.78  & 0.88  \\
  ~ & ~       & DRY & Clothes Dryer       & 0.98  & 0.85  \\
  ~ & ~       & DSH & Dishwasher          & 0.84  & 0.90  \\
  \hline

  4 & REDD    & FRG & Refrigerator        & 0.91  & 0.89  \\
  ~ & House 3 & DSH & Dishwasher          & 0.83  & 0.90  \\
  ~ & ~       & DRY & Clothes Dryer       & 0.83  & 0.99  \\
  ~ & ~       & MIC & Microwave           & 0.59  & 0.87  \\
  ~ & ~       & BTH & Bathroom GFI        & 0.68  & 0.91  \\
  ~ & ~       & FRN & Furnace             & 0.66  & 0.76  \\
  ~ & ~       & SMK & Smoke Detector      & 0.16  & 0.64  \\
  \hline

  5 & REDD    & LTE & Lights              & 0.75  & 0.73  \\
  ~ & House 4 & KTC & Kitchen Outlets     & 0.64  & 0.80  \\
  ~ & ~       & DRY & Clothes Dryer       & 0.67  & 0.65  \\
  ~ & ~       & STV & Stove               & 0.64  & 0.66  \\
  ~ & ~       & ARC & Air Conditioner     & 0.57  & 0.53  \\
  ~ & ~       & SMK & Smoke Detector      &-0.32  & 0.79  \\
  ~ & ~       & DSH & Dishwasher          & 0.70  & 0.88  \\
  ~ & ~       & BTH & Bathroom GFI        & 0.82  & 0.89  \\
  \hline

  6 & REDD    & MIC & Microwave           & 0.38  & 0.55  \\
  ~ & House 5 & LTE & Lighting            & 0.77  & 0.84  \\
  ~ & ~       & UKN & Unknown Circuit     & 0.61  & 0.67  \\
  ~ & ~       & SBP & Sub-Panel           & 0.62  & 0.72  \\
  ~ & ~       & HEA & Heater              & 0.91  & 0.93  \\
  ~ & ~       & DIF & Unmetered           & 0.75  & 0.96  \\
  \hline

  7 & REDD    & ELE & Electronics         & 0.70  & 0.70  \\
  ~ & House 6 & BTH & Bathroom GFI        & 0.53  & 0.56  \\
  ~ & ~       & FRG & Refrigerator        & 0.64  & 0.77  \\
  ~ & ~       & UKN & Unknown Circuit     & 0.68  & 0.73  \\
  ~ & ~       & LTE & Lighting            & 0.64  & 0.91  \\
  ~ & ~       & ARC & Air Conditioner     & 0.73  & 0.98  \\
  ~ & ~       & DIF & Unmetered           & 0.51  & 0.70  \\
  \hline
  \end{tabular}
  \normalsize
  \label{tbl:allres}
  \vspace{-0.2 in}
\end{table}

In Experiment 2 we used  $n=7$ appliances (OVN, RFG, DSH, MIC, DRY, BTH, and DIF) from house 1 of the REDD dataset~\cite{kolter2011}, with $m=13$ states.  We down-sampled the data by a factor 20 to obtain the samples at 1-minute intervals.  The AC values are given in Table~\ref{tbl:allres} and the overall ACC in Table~\ref{tbl:overallres}. Again, ALIP performs better than IP on all appliances individually, as well as overall.

In Experiment 3 we used  $n=6$ appliances (KTC, LTE, STV, MIC, DRY, and DSH) from house 2 in~\cite{kolter2011}, with $m=17$ states.  We downsampled the data by a factor 5 to obtain the samples at 15-second intervals. The AC values are given in Table~\ref{tbl:allres}. As seen in the results, ALIP performs significantly better than IP on all appliances except DRY. This is because DRY has a state with a high VA rating, whose occurrence is infrequent, and the median filter has filtered out some of its occurrences. As a result, the accuracy of ALIP gets reduced compared to IP on DRY. Nonetheless, the accuracy of ALIP on DRY is still acceptable and its overall accuracy is significantly better than that of IP, as seen in Table~\ref{tbl:overallres}.

In Experiment 4 we used  $n=7$ appliances (KTC, LTE, STV, MIC, DRY, and DSH) from house 3 in~\cite{kolter2011}, with $m=20$ states.  We downsampled the data by a factor 10 to obtain the samples at 30-second intervals.  The AC values are given in Table~\ref{tbl:allres}, where we see that ALIP performs significantly better than IP on all appliances except FRG, where it performs similarly due to the reasons discussed in Experiment 1. Overall, again, ALIP outperforms IP by a significant margin as Seen in Table~\ref{tbl:overallres}.

In Experiment 5 we used  $n=8$ appliances (LTE, KTC, DRY, STV, ARC, SMK, DSH, and BTH) from house 4 in~\cite{kolter2011}, with $m=20$ states. The data was downsampled factor 10 to obtain the samples at 30-second intervals. As seen in Table~\ref{tbl:allres}, IP performs slightly better than ALIP on LTE, DRY and ARC, but the difference is small, within 4\%. However, on other appliances ALIP performs better than IP, and in most cases by a significant margin. Note that IP yields a negative AC value for SMK, which means that it produces false positives more often than true positives.  Overall, ALIP has a significant advantage over IP, as seen in Table~\ref{tbl:overallres}.

\begin{table}[!hb]
\vspace{-0.1 in}
  \centering
  \caption{ACC results (incl. published comparison)}
  \footnotesize
  \begin{tabular}{c @{}| c@{}| c@{}| c@{}|c@{}|c@{}|c@{}|c@{}}
  \hline
  \hline
  ~& Exp.~1&Exp.~2&Exp.~3 & Exp.~4& Exp.~5& Exp.~6& Exp.~7\\
\hline
IP &0.92 & 0.51&0.70& 0.64& 0.43& 0.57&0.46\\
\hline
ALIP&$\bm{0.96}$ & 0.76&0.82&$\bm{0.87}$&$\bm{0.76}$
& $\bm{0.83}$& $\bm{0.79}$\\
\hline
\cite{kolter2011} & -- & 0.47 & 0.51 & 0.33 & 0.52 &   --   & 0.56\\
\hline
\cite{johnson13} & -- & $\bm{0.82}$ & $\bm{0.85}$ & 0.82 &  --    &   --   & 0.78\\
\hline
  \end{tabular}
  \normalsize
  \label{tbl:overallres}
  \vspace{-0.2 in}
\end{table}
\begin{table}[!hb]
\vspace{-0.1 in}
  \centering
  \caption{Average execution time per data sample (milliseconds)}
  \footnotesize
  \begin{tabular}{ c@{}| c@{}| c@{}| c@{}|c@{}|c@{}|c@{}|c@{}}
  \hline
  \hline
  ~&Exp.~1& Exp.~2&Exp.~3 & Exp.~4& Exp.~5& Exp.~6& Exp.~7\\
\hline
IP &14.2& 15.3 & 15.1 & 14.7& 16.0& 17.0&17.3\\
\hline
ALIP&15.4& 16.7 & 17.0 & 16.9&18.5 & 17.0& 18.8\\
\hline
  \end{tabular}
  \normalsize
  \label{tbl:exetime}
  \vspace{-0.1 in}
\end{table}

In Experiment 6 we used  $n=6$ appliances (MIC, LTE, UKN, SBP, HEA, and DIF) from house 5 in~\cite{kolter2011}, with $m=24$ states. The data were downsampled by a factor 10 to obtain the samples at 30-second intervals. Here, ALIP outperforms IP on all appliances, as well as in overall accuracy, as seen in Tables~\ref{tbl:allres} and~\ref{tbl:overallres}, respectively.

Finally, in Experiment 7 we used  $n=7$ appliances (ELE, BTH, FRG, UKN, LTE, ARC, and DIF) from house 6 in~\cite{kolter2011}, with $m=20$ states.  The data were downsampled by a factor 10 to obtain the samples at 30-second intervals. Again, ALIP outperforms IP consistently on all appliances, as well as in overall accuracy, as seen in Tables~\ref{tbl:allres} and~\ref{tbl:overallres}, respectively. 

In Table~\ref{tbl:overallres} we also include the published ACC results of two state-of-the-art machine learning-based disaggregation approaches,~\cite{kolter2011} and~\cite{johnson13}, on the REDD dataset. Note that~\cite{kolter2011} and~\cite{johnson13} did not report the results for all the houses. Although these methods used different downsampling rates and there is some uncertainty about the  processing of data prior to testing, the comparison gives us a feeling for how the proposed ALIP would compare against machine learning-based disaggregation. We note that against~\cite{kolter2011}, ALIP scores on average 0.32 better. Against~\cite{johnson13}, it scores on average 0.03 better on houses 3 and 6, and on average 0.04 lower on houses 1 and 2. Based on these results, we conclude that the proposed ALIP is competitive in terms of accuracy with the state-of-the-art machine learning-based disaggregation approaches.

A final word on complexity. The proposed ALIP approach incorporates several additional processing steps compared to the conventional IP-based disaggregation. Hence, its computational complexity is slightly higher. In Table~\ref{tbl:exetime} we list the average execution time (in milliseconds) per sample of IP and ALIP disaggregators for Experiments~1--7, which were obtained in Matlab 2015a using \texttt{intlinprog} and \texttt{linprog} (with default settings) on an Intel(R) Core(TM) i7-4770 CPU@3.40 GHz processor with 16 GB RAM. As seen in the table, ALIP is only slightly slower than IP, and both disaggregators take less than 20~ms per data sample.

\section{Conclusion}
\label{sec:sec6}
Integer programming (IP) provides a natural way to solve the load disaggregation problem, by trying to determine which appliance states are active at any given time. However, previous IP-based disaggregation is shown to run into problems on real data due to issues related to transient readings and in cases when some states are binary combinations of other states. We proposed an aided linear IP (ALIP) approach for disaggregation that overcomes many of the shortcomings of the previous IP-based approach.
Experimental results demonstrate significant accuracy advantage of ALIP over the previous IP-based disaggregation method, as well as competitive performance against two state-of-the art machine learning-based disaggregation approaches.


\end{document}